\documentclass[letterpaper, 10 pt, conference]{ieeeconf}  % Comment this line out if you need a4paper
\IEEEoverridecommandlockouts 
\usepackage{subfigure}
\usepackage{color}
\usepackage{graphicx}    
\usepackage{float}
\usepackage{multirow}
\usepackage{algorithm}
\usepackage{algpseudocode}
\usepackage{amsmath}
\usepackage[top=2cm, bottom=2cm, left=2cm, right=2cm]{geometry}
\usepackage{algorithmicx}
\usepackage{algpseudocode}
\usepackage{hyperref}
\usepackage{threeparttable}
\usepackage{booktabs}

\usepackage{indentfirst}
\usepackage{amsmath}
\usepackage{cite}

\usepackage{color}
\usepackage{soul}
\soulregister\cite7
\soulregister\ref7

\title{\LARGE \bf
TransSC: Transformer-based Shape Completion for Grasp Evaluation
}

\author{Wenkai Chen$^{1*}$,  Hongzhuo Liang$^{1}$, Zhaopeng Chen$^{2}$, Fuchun Sun$^{3}$ and Jianwei Zhang$^{1}$% <-this % stops a space
\thanks{$^{1}$Technical Aspects of Multimodal Systems (TAMS), Department of Informatics, Universit\"{a}t Hamburg}% <-this % stops a s
\thanks{$^{2}$Agile Robots AG}%
\thanks{$^{3}$Beijing National Research Center for Information Science and Technology (BNRist), State Key Lab on Intelligent Technology and Systems, Department of Computer Science and Technology, Tsinghua University}
\thanks{*Corresponding author to provide e-mail: wchen@informatik.uni-hamburg.de}%
}

\begin{document}
\maketitle
\thispagestyle{empty}
\pagestyle{empty}

%%%%%%%%%%%%%%%%%%%%%%%%%%%%%%%%%%%%%%%%%%%%%%%%%%%%%%%%%%%%%%%%%%%%%%%%%%%%%%%%
\begin{abstract}
Currently, robotic grasping methods based on sparse partial point clouds have attained a great grasping performance on various objects while they often generate wrong grasping candidates due to the lack of geometric information on the object. In this work, we propose a novel and robust shape completion model (TransSC).  This model has a transformer-based encoder to explore more point-wise features and a manifold-based decoder to exploit more object details using a partial point cloud as input.
 Quantitative experiments verify the effectiveness of the proposed shape completion network and demonstrate it outperforms existing methods. Besides, TransSC is integrated into a grasp evaluation network to generate a set of grasp candidates. The simulation experiment shows that TransSC improves the grasping generation result compared to the existing shape completion baselines.  Furthermore, our robotic experiment shows that with TransSC the robot is more successful in grasping objects that are randomly placed on a support surface.
\end{abstract}

%%%%%%%%%%%%%%%%%%%%%%%%%%%%%%%%%%%%%%%%%%%%%%%%%%%%%%%%%%%%%%%%%%%%%%%%%%%%%%%%
\section{Introduction}
Robotic grasping evaluation is a challenging task due to incomplete geometric information from single-view visual sensor data~\cite{varley2015generating}. Many probabilistic grasp planning models have been proposed to address this problem, such as Motel Carlo, Gaussian Process and uncertainty analysis~\cite{lundell2019robust, tosun2020robotic, gualtierirobotic}. However, these analytic methods are always computationally expensive. With the development of deep learning techniques, data-driven grasp detection methods have shown great potential~\cite{breyer2021volumetric,wu2020grasp,ten2017grasp,liang2019pointnetgpd} to solve this problem. They generate lots of grasp candidates and estimate the corresponding grasp quality, resulting in a better grasp performance and generalization. But as most of these methods still rely on original sensor input like 2D (image) and 2.5D (depth map), there exists a physical grasping defect when the gripper interacts with real object surfaces or edges because of the incomplete pixel-wise and point-wise representations. 

\begin{figure}[!ht]
\centering
\includegraphics[width=0.48\textwidth]{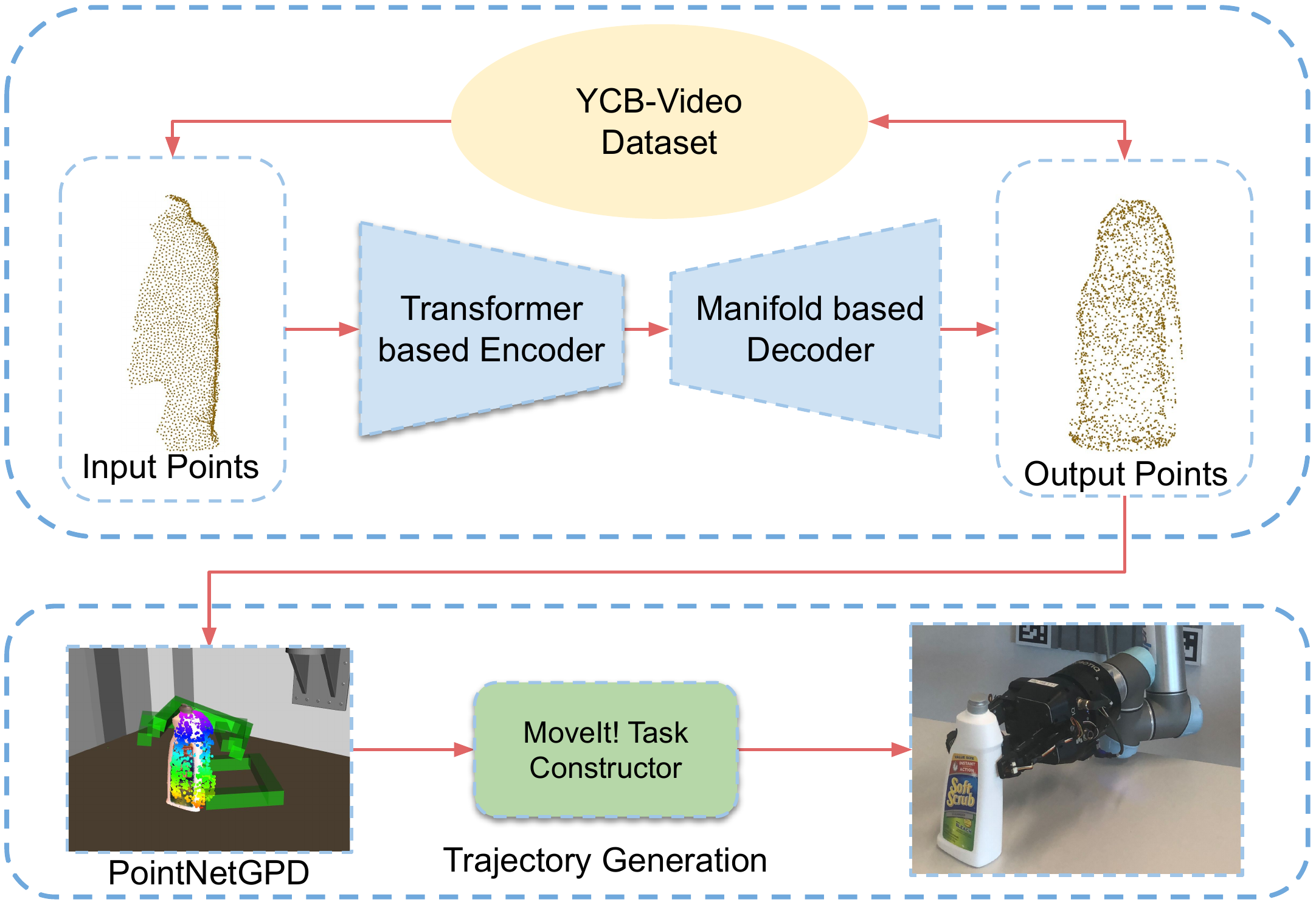}
\caption{Overview of our shape completion based grasp pipeline. 
The upper line is the shape completion module. In this module, a partial point cloud $\zeta_p$ with \emph{n} points is first input into a transformer-based encoder to extract point-wise and self-attention features, which outputs a latent vector with \emph{m} dimensions. Then, the latent vector is concatenated with another latent feature from a flat/spatial point seed generator to predict multiple spatial surfaces in the manifold-based decoder. Finally these surfaces are montaging into a complete point cloud $\zeta_c$. The bottom line is the grasp evaluation module, the complete point cloud $\zeta_c$ is the input of our grasp detection pipeline PointNetGPD to compute the grasp quality $\mathcal{Q}_i$. The grasp with the highest score $\mathcal{G}_{best}$ will be send to calculate collision free trajectory and executed in a real robot experiment.}
\label{Overview}
\vspace{-0.5em}
\end{figure}

To cope with this limitation, the missing geometric and semantic information of the object needs to be restored or repaired to generate a better grasping interaction. Additional sensor input such as tactile sensor is introduced to supplement original vision sensing~\cite{watkins2019multi}. However, object uncertainty still exists and extra sensor interference with the object will directly affect the final grasping result. Another strategy is to use shape completion to infer the original object shape while traditional grasping-based shape completion methods use a high-resolution voxelized grid as object representation \cite{varley2017shape,lundell2019robust,lundell2020beyond}, causing a high memory cost and information loss since the sparsity of the sensory input. To avoid extra sensor cost and obtain complete object information, a novel transformer-based shape completion module is proposed in this work based on an original sparse point cloud. Compared with the traditional convolutional network layer, the transformer has achieved state-of-the-art results in the visual recognition and segmentation~\cite{srinivas2021bottleneck,guo2020pct}, which enables our shape completion module to achieve a better performance.

As illustrated in Fig.~\ref{Overview}, we present a novel grasping pipeline that uses a sparse point cloud to execute the grasp directly without converting it into voxel grids during the shape completion process and transforming it into mesh in the grasp planning process. The pipeline consists of two sub-modules: The transformer-based shape completion module and the grasp evaluation module. In the first module, a non-synthetic partial point cloud dataset based on a YCB object was constructed. Not cropping the object randomly or viewing the object in a physical simulator, our dataset contains lots of real cameras and environmental noise, which guarantees an improved grasping interaction in a real robot environment. Based on this dataset, we propose a novel encoder-decoder point cloud completion network architecture (TransSC), which outperforms some representative baselines in different evaluation metrics. In the second module, our previous work~\cite{liang2019pointnetgpd} is referred. We use PointNet to obtain feature representation of the repaired point cloud and build a grasp detection network to generate and evaluate a set of grasp candidates. The grasp with the highest score will be executed in the real robot experiment. The proposed pipeline is validated in a simulation experiment and a robotic experiment, which all demonstrate our shape completion pipeline can improve grasping performance significantly.

The main contributions of this paper can be summarized as:
\begin{itemize}
\item A large-scale non-synthetic partial point cloud dataset is constructed based on the YCB-Video dataset. As the dataset is based on 3D point cloud data captured by a real RGB-D camera, the noise that comes from it will facilitate the generalization of our work. 
\item A novel point cloud completion network Transformer-based Shape Completion (TransSC) is proposed. The transformer-based encoder and manifold-based decoder are introduced into the shape completion task for a better shape completion performance.
\item Combining our previous work PointNetGPD for grasp evaluation and the Moveit! Task Constructor for motion planning, we demonstrate a robust grasp planning pipeline that using the shape completion result as input could get a better grasp planning result compared to the single view and no shape completion work.
\end{itemize}

%%%%%%%%%%%%%%%%%%%%%%%%%%%%%%%%%%%%%%%%%%%%%%%%%%%%%%%%%%%%%%%%%%%%%%%%%%%%%%
\section{RELATED WORK}
\textbf{Deep Visual Robotic Grasping}
With the development of deep learning, many methods for deep visual grasping have been proposed. Borrowing from the ideas of 2D object recognition, monocular camera images were firstly used to predict the probability that the input grasps were successful \cite{levine2018learning}. In \cite{chu2018real} and \cite{tosun2020robotic}, a single RGB-D image of the target object was used to generate a 6D-pose grasp and effective end-effector trajectories. However, these works are not suitable to deal with sparse 3D object information and spatial grasps. Compared with the 2D feature representations from images, 3D voxel or point cloud data could provide robotic grasping with more semantic and spatial information.  Given a synthetic grasp dataset, \cite{breyer2021volumetric} transformed scanning 3D object information into Truncated Signed Distance Function (TSDF) representations and passed them into a Volumetric Grasping Network (VGN) to directly output grasp quality, gripper orientation and gripper width at each voxel. \cite{wu2020grasp} designed a special grasp proposal module that defines anchors of grasp centers and related 3D grid corners to predict a set of 6D grasps from a partial point cloud. \cite{ten2017grasp} used hand-crafted projection features on a normalized point cloud to construct a CNN-based grasp quality evaluation model. In our previous work \cite{liang2019pointnetgpd}, we used PointNet \cite{qi2017pointnet} to extract raw point cloud features and built a grasp evaluation network, which achieves a great performance in robotic grasping experiments.

\textbf{Grasp-based Shape Completion}
For robotic grasping, the key challenge is to recognize objects in 3D space and avoid potential perception uncertainty. When the RGB-D camera captures an object from a particular viewpoint, the 3D information of the object is incomplete, which means a lot of semantic and spatial information is missing. This will affect the quality of later generated grasping and cause wrong grasping poses. 

Recently, some researchers proposed to use shape completion to enable robotic grasping. In \cite{varley2017shape}, the observed object from 2.5D range sensors was firstly converted to occupancy voxel grid data. Then the voxelized data were input into a CNN network and formed a high-resolution voxel output. Furthermore, the completion result was transformed into mesh and then loaded into Graspit!~\cite{miller2004graspit} to generate a grasp.~\cite{lundell2019robust} used dropout layers to modify the network, which enabled the prediction of shape samples at run-time. Meanwhile, Monte Carlo Sampling and probabilistic grasp planning were used to generate grasp candidates. As traditional analytic grasping methods are computationally expensive, \cite{lundell2020beyond} combined the shape completion of a voxel grid and a data-driven grasping planning strategy (GQCNN) \cite{mahler2017dex}  to propose a structure called FC-GQCNN, where synthetic object shapes were obtained from a top-down physics simulator and grasps were generated from depth images. 

In conclusion, traditional grasp shape completion methods mainly voxelized the 2.5D data into occupancy grids or distance fields to train a convolutional network. However, these high-resolution voxel grids will entail a high memory cost. Moreover, detailed semantic information is often lost as an artifact of discretization, which causes meaningful geometric features of objects not to be learned from the neural network. To obtain more complete geometric features and retain original object information, a transformer-based shape completion module is introduced in our proposed method. Without converting the observed partial point cloud into the voxel grid and mesh, our completion method outputs a repaired point cloud at arbitrary resolution and outperforms existing methods. Furthermore, PointNet\cite{qi2017pointnet} is introduced for the representation learning of the repaired point cloud and a grasp evaluation network is constructed to generate grasp candidates. Therefore, our grasp evaluation framework also achieves a better grasping performance than original framework without shape completion.

\begin{figure*}[!t]
\centering
\includegraphics[width=1.0\linewidth,height=5.5cm]{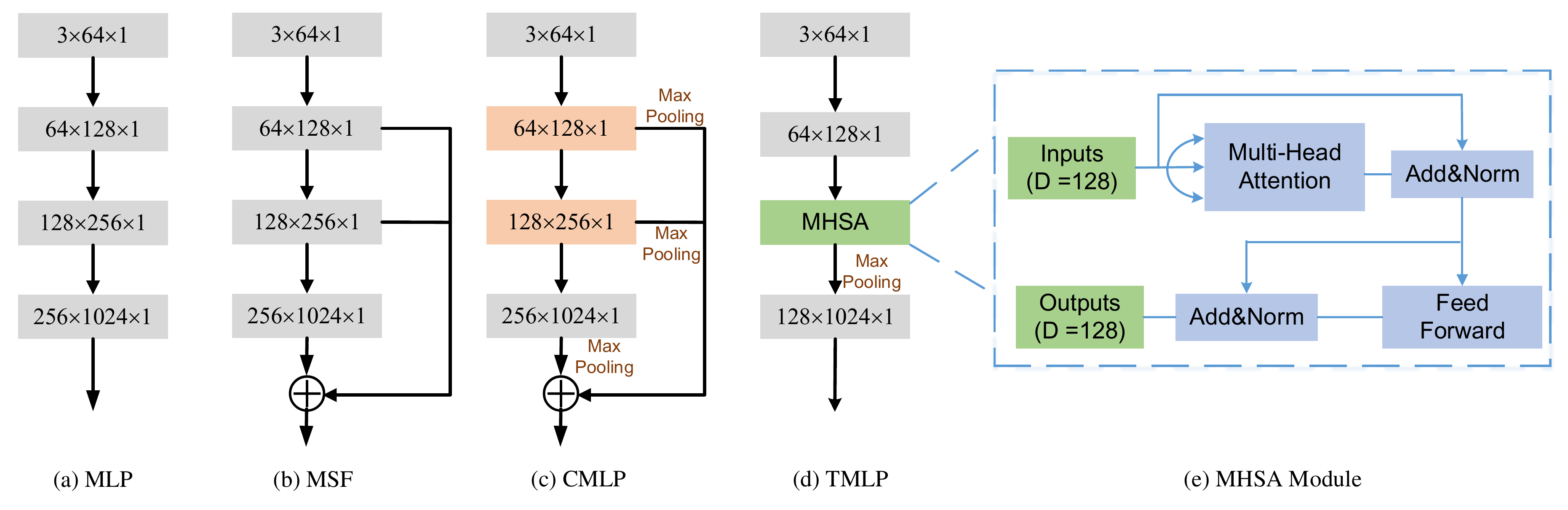}
\caption{Illustration of various encoder structures for point cloud completion. (a) is a simple multiple-layer perception (MLP) structure.  (b) is a multi-scale fusion (MSF) module, which can fuse features from different layers directly. (c) is concatenated multiple layer perception (CMLP), it also can concatenate multi-dimensional latent features while the max pooling operation is used to extract latent features further. (d) shows our Transformer-based multiple layer perception (TMLP) module, which integrates the Multi-head Self-attention (MHSA) module into the MLP structure. (e) depicts the architecture of the MHSA module.}  
\label{Encoder}
\vspace{-0.5em}
\end{figure*}
%%%%%%%%%%%%%%%%%%%%%%%%%%%%%%%%%%%%%%%%%%%%%%%%%%%%%%%%%%%%%%%%%%%%%%%%%%%%%%
\section{PROBLEM FORMULATION}
We consider a setup consisting of a robotic arm with parallel-jaw grippers, an RGB-D camera and an object to be grasped. Also, we assume that the RGB-D camera could capture the depth map of an object and convert it to a 2.5D partial point cloud $\mathcal{P} \in \mathcal{R}^{N\times3}$. For simplicity, all spatial quantities are in camera coordinates.

Given a gripper configuration $\mathcal{C}$ and camera observation $\mathcal{P}$, our goal is firstly to learn an encoder-decoder point cloud completion network, which could repair an observed 2.5D partial point cloud $\mathcal{P} \in \mathcal{R}^{N\times3}$, turning it into a complete 3D point cloud $\mathcal{P}_c \in \mathcal{R}^{N\times3}$. After that, a grasp evaluation network based on $\mathcal{P}_c$ is trained to predict a set of grasp candidates $\mathcal{G}_i$ and compute relative grasp quality $\mathcal{Q}_i$. The grasp with the highest score $\mathcal{G}_{best}$ will be executed in the real robot experiment.

%%%%%%%%%%%%%%%%%%%%%%%%%%%%%%%%%%%%%%%%%%%%%%%%%%%%%%%%%%%%%%%%%%%%%%%%%%%%%%
\section{Robotic Grasping Evaluation Via Shape Completion and Grasp Detection}

\subsection{Dataset Construction}
Traditional shape completion methods use synthetic CAD models from the ShapeNet~\cite{yi2016scalable} or ModelNet~\cite{wu20153d} datasets to generate partial and corresponding complete point cloud data, while these synthetic data contain little noise from the camera and robotic environment. In order to simulate real point cloud data distribution, we summarize a shape completion dataset from the YCB-Video Dataset~\cite{xiang2017posecnn}. Non-synthetic RGB-D video images ($\sim$ 133,827 frames) in the YCB-Video Dataset are firstly chosen, while most of them vary insignificantly. 
Thus, a pre-processed image dataset is obtained by reducing every 5 frames. Meanwhile, to cover distinguishable shapes with different levels of detail, 18 objects are also chosen in the YCB-Video dataset. In this work, the ground-truth point cloud of 18 objects is created by the farthest point sampling (FPS) 2048 points on each object model. Not randomly sampling or cropping complete point clouds on the unit sphere to get partial point clouds, RGB-D images and related object label images in the pre-processed dataset are loaded to compute the matching partial point clouds using related camera intrinsic parameters. To approximate the distribution of point cloud data of real objects and retain the semantic information, a large number of cameras and environmental noise data are kept on, though a certain radius is used to remove partial outliers. For the convenience of network training, the partial point clouds are also unified into the size of 2048 points by FPS or replicating points. To enable an accurate comparison with existing baselines, the canonical center of the partial point cloud of each object is transformed into the canonical center of the ground-truth point cloud using pose information. Finally, more than 70,000 partial point clouds are collected in our dataset. Compared to other synthetic point cloud datasets, our dataset also does well at preserving the real point cloud distribution of occluded objects.

\subsection{Transformer-based Encoder Module}
As shown in Fig.~\ref{Encoder}, we compare our proposed encoder module with several common competitive methods. Multi-layer Perception (MLP) is a simple baseline architecture to extract point features. This method maps each point into different dimensions and extracts the maximum value from the final $K$ dimensions to formulate a latent vector. A simple generalization for MLP is to combine semantic features from a low-level dimension with those of a high-level dimension. The MSF (Multi-scale Fusion)~\cite{kuang2020voxel} module inflates the dimension of the latent vector from 1024 to 1408 to obtain semantic features from different dimensions. To improve the performance of the feature extractor, L-GAN~\cite{achlioptas2018learning} proposed to use a Maxpooling layer appropriately. Concatenated Multiple Layer Perception (CMLP)~\cite{huang2020pf} maxpools the output of the last $k$ layers to guarantee that multi-scale feature vectors are concatenated directly. An overview of our proposed Transformer-based multi-layer perception (TMLP) module is shown in Fig.~\ref{Encoder}(d). Without an extra skip connection structure and a maxpooling operation from different layers, the Multi-head Self-attention (MHSA)~\cite{vaswani2017attention} module is introduced to replace the traditional convolutional layer [$128\times256\times1$].

\begin{figure*}[t]
\includegraphics[width=1.0\textwidth]{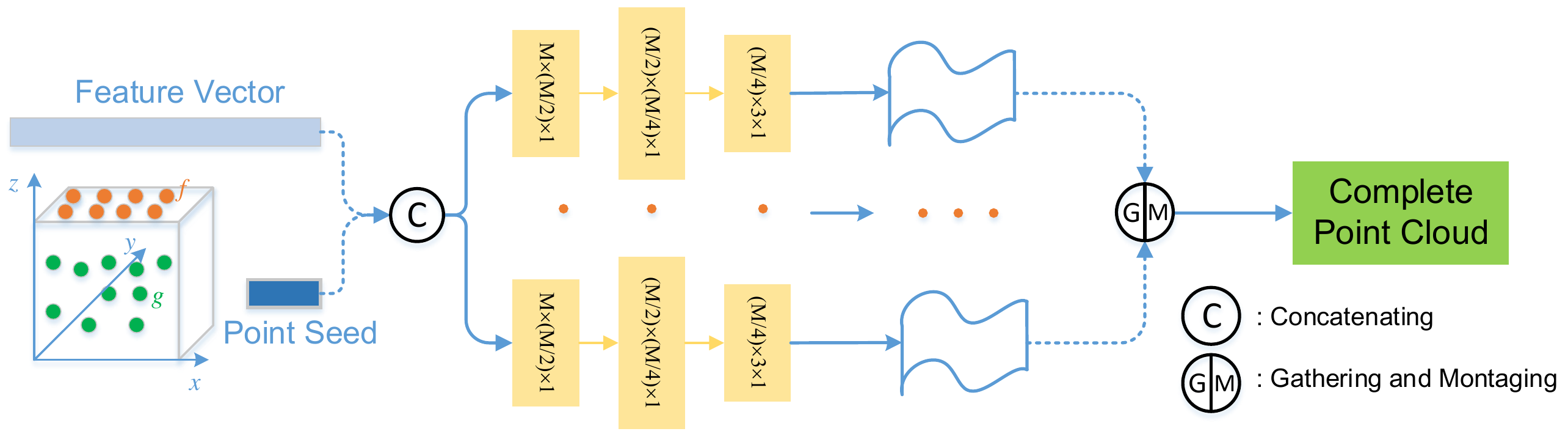}
\caption{Illustration of the decoder structure for point cloud completion. The feature vector with \emph{m} dimensions from the encoder is firstly concatenated with latent feature from a special point seed generator \emph{f} or \emph{g}. Then three convolutional layers as the backbone are used to extract features and form different manifold-based surfaces, respectively. Finally, these surfaces are gathering and montaging into a complete point cloud.}   
\label{Decoder}
\vspace{-0.5em}
\end{figure*}

MHSA aims to transform (encode) the input point feature into a new feature space, which contains point-wise and self-attention features. Fig.~\ref{Encoder}(e) shows a simple MHSA architecture used in TMLP, which includes two sub-layers. In our first layer, the multi-head number is set to 8 and the input feature dimension for each point is 128. Unlike natural language processing (NLP) problems, the 128-dimensional feature vector $\mathcal{A}_{in} \in \mathcal{R}^{2048\times128}$ will enter into the multi-head attention module directly without positional encoding. This is because each point in the point cloud has its unique $x-y-z$ coordinates. The output feature $\mathcal{Z}$ is formed by concatenating the attention of each attention head. A residual structure is also used to add and normalize the output feature $\mathcal{Z}$ with $\mathcal{A}_{in}$. This process can be formulated as follows:
\begin{equation}
    \mathcal{A}_{i} = SA_i(\mathcal{A}_{in})\quad  i=1,2,...,8
\end{equation}
\begin{equation}
    \mathcal{Z} = concat(\mathcal{A}_{1},\mathcal{A}_{2},...,\mathcal{A}_{8})*W_0
\end{equation}
\begin{equation}
    \mathcal{A}_{out} = Norm(\mathcal{A}_{in} + \mathcal{Z})
\end{equation}

where $SA_i$ represents the $i$-th self-attention layer, each has the same output dimension size with input feature vector $\mathcal{A}_{in}$, and $W_0$ is the weight of the linear layer. $\mathcal{A}_{out}$ represents the output point-wise features of the first sub-layer.

The second sub-layer is called Feed-forward module, which is a fully connected network. Point-wise features $\mathcal{A}_{out}$ are processed through two linear transformations and one ReLU activation. Furthermore, a residual network is also used to fuse and normalize the output features. Finally, we can get the MHSA module output $\mathcal{FF}_{out} \in \mathcal{R}^{2048\times128}$ as: 

\begin{equation}
    \mathcal{FF} = ReLU(\mathcal{A}_{out} * W_1 + b_1) *W_2 + b_2
\end{equation}

\begin{equation}
    \mathcal{FF}_{out} = Norm(\mathcal{A}_{out} + \mathcal{FF})
\end{equation}

where $W_1$, $W_2$ and $b_1$, $b_2$ represent the weight and bias value of the corresponding linear transformation, respectively.

\subsection{Manifold-based Decoder Module}
Inspired by the AtlasNet~\cite{groueix1802atlasnet}, a manifold-based decoder module is designed to predict a complete point cloud from partial point cloud features. As shown in Fig.~\ref{Decoder}, a complete point cloud could can be assumed that it consists of multiple sub-surfaces. Therefore, we only concentrate on obtaining each sub-surface, then we gather them and make appropriate montage to form the final complete point cloud. To get each sub-surface, a point seed generator is used to concatenate with global feature vector $\mathcal{P}_{g} \in \mathcal{R}^{2048\times1024}$ output from the encoder, where point initialization values are computed from a flat $(f)$ or spatial $(g)$ sampler. As the coordinate values of the ground-truth point cloud are limited to between [-1, 1], point initialization values are also limited in this range. After that, the concatenated feature vector $\mathcal{P}_{concat} \in \mathcal{R}^{2048\times M} (M = 1026$\ $or$\ $1027)$  is input into $K$ convolutional layers, where all sampled 2D or 3D points will be mapped to 3D points on each sub-surface. In our decoder, the sub-surface number is set to 16. Unlike other voxel-based shape completion methods, our decoder module achieves an arbitrary resolution for the final completion results.

\textbf{Evaluation Metrics}
To evaluate our shape completion results, we used two permutation-invariant metrics called Chamfer Distance (CD) and Earth Mover's Distance (EMD) as our evaluation goal~\cite{fan2017point}. Given two arbitrary point clouds $S_1$ and $S_2$, CD measures the average distance between each point in one point cloud to its nearest point coordinates in the other point cloud. 
\begin{equation}
   d_{CD}\left(S_{1}, S_{2}\right)=\frac{1}{S_{1}} \sum_{x \in S_{1}} \min _{y \in S_{2}}\|x-y\|_{2}^{2}+\frac{1}{S_{2}} \sum_{y \in S_{2}} \min _{x \in S_{1}}\|y-x\|_{2}^{2} 
\end{equation}

 While Earth Mover's Distance considers two equal point sets $S_1$ and $S_2$ and is defined as:
\begin{equation}
  d_{EMD}\left(S_{1}, S_{2}\right)=\min _{\emptyset: S_{1} \rightarrow S_{2}} \frac{1}{S_{1}} \sum_{x \in S_{1}}\|x-\emptyset(x)\|_{2}  
\end{equation}

CD has been widely used in most shape completion tasks because it is efficient to compute. However, EMD is chosen as our completion loss because CD is blind to some visual inferiority and ignores details easily~\cite{achlioptas2018learning}. With ${\emptyset: S_{1} \rightarrow S_{2}}$ being bijective, EMD could solve the assignment and transformation problem in which one point cloud is mapped into another. 

\begin{table*}[htb!]

\centering
\caption{Comparison of Earth Mover's Distance in different point cloud completion models}
\resizebox{1\textwidth}{!}{
\label{EMD_distance}
\begin{tabular}{l|lllllllll}
\hline
\textbf{Model} & \textbf{\begin{tabular}[c]{@{}l@{}}cracker \\ box\end{tabular}} & \textbf{banana} & \textbf{\begin{tabular}[c]{@{}l@{}}pitcher\\ base\end{tabular}} & \textbf{\begin{tabular}[c]{@{}l@{}}bleach\\ cleanser\end{tabular}} & \textbf{bowl} & \textbf{mug} & \textbf{\begin{tabular}[c]{@{}l@{}}power \\ drill\end{tabular}} & \textbf{scissors} & \textbf{average}                      \\ \hline
\textbf{Oracle}        & 3.4                 & 1.7                    & 4.6                  & 2.9         & 1.9                  & 2.0    &3.8   &1.5  &2.7                                         \\
\textbf{AtlasNet \cite{groueix1802atlasnet}}        & 9.7                 & 4.9                    & 10.5                  & 10.0         & 8.8                  & 5.3    &15.0   &5.2 &8.7                                          \\
\textbf{MSN (fusion) \cite{liu2020morphing}}                    & 10.7                 & 4.6                    & 12.4                  & 14.0         & 11.5                  & 12.9    &23.4   &5.3  &11.8                              \\
\textbf{MSN (vanilla) \cite{liu2020morphing}}  & 11.0                 & \textbf{3.8}                    & 9.3                  & 8.3         & 10.2                  & 3.9    &5.9   &\textbf{3.4}   &7.0                       \\
\textbf{Our (flat)}                 & \textbf{8.5}                 & 3.9                    & 9.4                  & 6.7         & 6.0                  & \textbf{3.7}    &\textbf{5.2}   &4.1  &\textbf{5.9} \\
\textbf{Our (spatial)}                 & 10.1                 & 4.4                    & \textbf{8.4}                  & \textbf{5.8}         & \textbf{5.6}                  & \textbf{3.7}    &7.0   &3.9 &6.1
\\ \hline
\end{tabular}}
\end{table*}

\begin{table*}[htb!]
\centering
\caption{Comparison of Chamfer Distance in different point cloud completion models}
\label{chamfer_distance}
\resizebox{1\textwidth}{!}{
\begin{tabular}{l|lllllllll}
\hline
\textbf{Model} & \textbf{\begin{tabular}[c]{@{}l@{}}cracker \\ box\end{tabular}} & \textbf{banana} & \textbf{\begin{tabular}[c]{@{}l@{}}pitcher\\ base\end{tabular}} & \textbf{\begin{tabular}[c]{@{}l@{}}bleach\\ cleanser\end{tabular}} & \textbf{bowl} & \textbf{mug} & \textbf{\begin{tabular}[c]{@{}l@{}}power \\ drill\end{tabular}} & \textbf{scissors} & \textbf{average}     \\\hline
\textbf{Oracle}        & 0.24                 & 0.52                    & 0.28                  & 0.12         & 0.10                  & 0.09    &0.13   &0.38     &0.23                                       \\
\textbf{AtlasNet \cite{groueix1802atlasnet}}        & 4.51                 & 0.87                    & 4.97                  & 5.61         & 4.21                  & 1.37    &6.18   &0.92       &3.58                                      \\
\textbf{MSN (fusion) \cite{liu2020morphing}}                    & 5.59                 & 1.25                    & 5.71                  & 2.77         & 10.81                  & 1.77    &8.34   &1.58     &4.73                             \\
\textbf{MSN (vanilla) \cite{liu2020morphing}}  & 6.01                 &\textbf{0.71}                    & 4.01                  & 4.68         & 7.51                  & 0.76    &1.28   &\textbf{0.38}    &3.17                        \\
\textbf{Our (flat)}                 &\textbf{3.28}                 & 0.92                    & 4.09                  & 1.50         & \textbf{2.55}                  & \textbf{0.66}    &\textbf{1.25}   &0.82      &\textbf{1.88}  \\
\textbf{Our (spatial)}                 & 5.81                 & 0.87                    & \textbf{3.19}                  & \textbf{1.20}         & 2.79                  & 0.69    &2.54   &0.66     &2.22                             
\\ \hline
\end{tabular}}
\end{table*}

\begin{table*}[htb!]
\centering
\caption{Comparison of EMD and CD from different encoder structures}
\resizebox{1.0\textwidth}{!}{
\label{Structure}

\begin{tabular}{l|l|l|l|l}
\hline
\multirow{1}{*}{\textbf{Earth Mover's Distance (EMD)}} & \multirow{1}{*}{\textbf{MLP}} & \multirow{1}{*}{\textbf{CMLP}} & \multirow{1}{*}{\textbf{MSF}} & \multirow{1}{*}{\textbf{TMLP}}
\\\hline
Mug                                            & 6.01              & 3.69            & 9.45         & {\textbf{3.69}}                                        \\
Bleach cleanser                                           & 10.51              & 8.10          & 11.70         & {\textbf{6.70}}                                    \\ \hline
\end{tabular}

\begin{tabular}{l|l|l|l|l}
\hline
\multirow{1}{*}{\textbf{Chamfer Distance (CD)}} & \multirow{1}{*}{\textbf{MLP}} & \multirow{1}{*}{\textbf{CMLP}} & \multirow{1}{*}{\textbf{MSF}} & \multirow{1}{*}{\textbf{TMLP}}
\\\hline
Mug                                            & 2.15              & {\textbf{0.65}}            & 13.80         & 0.66                                        \\
Bleach cleanser                                           & 6.88              & 2.63          & 13.89         & {\textbf{1.50}}                                    \\ \hline
\end{tabular}}
\vspace{-1.0em}
\end{table*}

\subsection{PointNetGPD: Grasping Detection Module}
Giving the complete point cloud from previous steps, we put the point cloud into a geometric-based grasp pose generation algorithm (GPG)~\cite{gpg}, which outputs a set of grasp proposals $\mathcal{G}_i$. We then transform $\mathcal{G}_i$ into a gripper coordinate system and use points inside the gripper as the input of PointNetGPD. The output grasp will then be sent to the MoveIt! Task Constructor~\cite{mtc_grasp} to plan a feasible trajectory for pick and place task.

\begin{figure}[t!]
    \centering
    \includegraphics[width=0.48\textwidth]{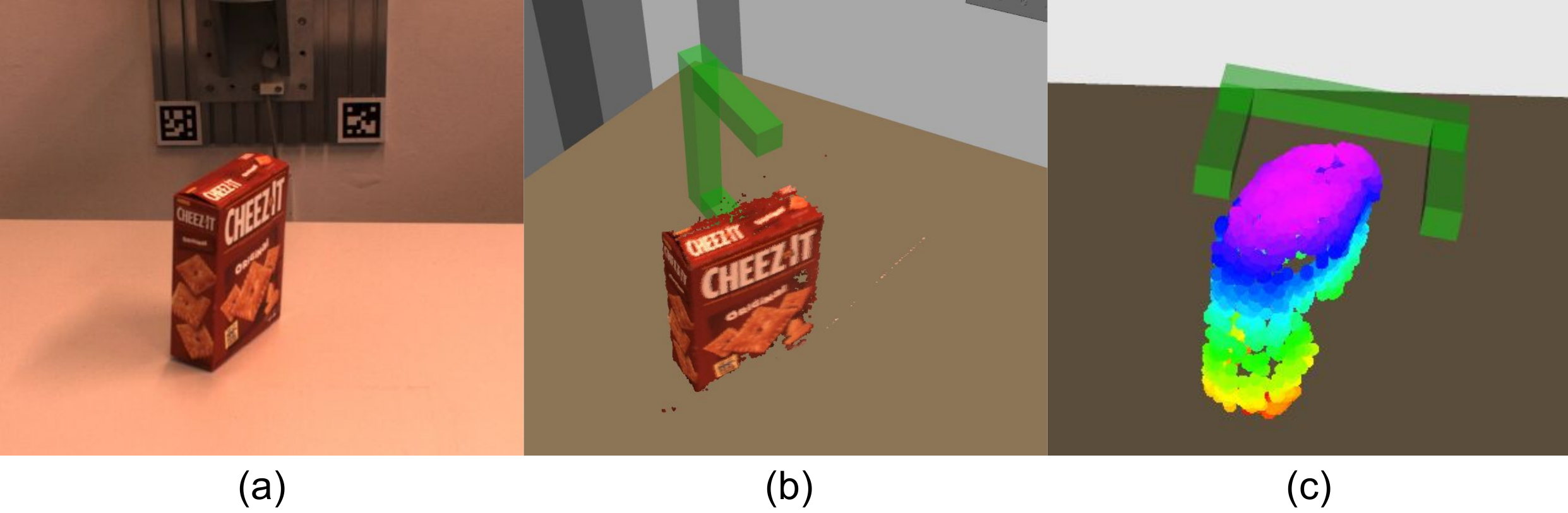}
    \caption{Comparison of grasp candidates generated using GPG~\cite{gpg}. (a) RGB image to show the example environment, (b) grasp generated with partial point cloud, (c) grasp generated with complete point cloud.}
    \label{fig:grasp_candidate}
\vspace{-0.5em}
\end{figure}

PointNetGPD is trained on a grasp dataset generated using reconstructed YCB object mesh and evaluates the input grasp quality. The grasp candidates in the grasp dataset are all collision-free with respect to the target object.
As a result, the grasp evaluation network assumes all the input grasp candidates are not colliding with the object. 
If the object has occlusion due to the camera viewpoint, current geometric-based grasp proposal algorithm will generate grasp candidates that collide with the object.
Thus, using a complete point cloud could ensure that the grasp candidate generation algorithm generates grasp sets that do not collide with the graspable objects.  Fig.~\ref{fig:grasp_candidate} shows the comparison of grasp generation result using GPG~\cite{gpg} with and without point cloud completion, where Fig.~\ref{fig:grasp_candidate}(b) shows a candidate generated using partial point cloud and Fig.~\ref{fig:grasp_candidate}(c) shows a grasp candidate generated using complete point cloud. We can see that the grasp in Fig.~\ref{fig:grasp_candidate}(b) has a collision with the real object while Fig.~\ref{fig:grasp_candidate}(c) avoids generating such that grasp. 

%%%%%%%%%%%%%%%%%%%%%%%%%%%%%%%%%%%%%%%%%%%%%%%%%%%%%%%%%%%%%%%%%%%%%%%%%%%%%%
\section{Experiments}
\subsection{Quantitative Evaluation of Proposed Shape Completion Network}
\textbf{Training and Implementation details}
To evaluate model performance and reduce training time, 8 categories of different objects in our dataset are chosen to train the shape completion model. The training set and validation set are split into 0.8:0.2. We implement our network on PyTorch. All the building modules are trained by using the Adam optimizer with an initial learning rate of 0.0001 and a batch size of 16. All the parameters of the network are initialized using a Gaussian sampler. Batch Normalization (BN) and ReLU activation units are all employed at the encoder and decoder module except the final tanh layer producing point coordinates, and Dropout operation is used in the MHSA module to suppress model overfitting.  
\subsubsection{Comparison with Existing Methods} 
In this subsection, we compare our method against several representative baselines that are also used for point cloud completion, including AtlasNet~\cite{groueix1802atlasnet} and MSN~\cite{liu2020morphing}. The Oracle method means that we randomly resample 2048 points from the original surface of different YCB objects. Corresponding EMD and CD distance between the resampled point cloud and the ground-truth point cloud provide an upper bound for the performance. Relative comparison results are shown in Table.~\ref{EMD_distance} and Table.~\ref{chamfer_distance}. Our method is developed into two models based on the different point seed generators $(f/g)$ in the decoder module. It can be seen that our method outperforms other methods in most objects on both EMD and CD distances. For the same completion loss, our (flat) model achieves an average of about 15\% improvement in terms of the EMD distance with respect to the latest MSN (vanilla) model. Since our dataset contains much noise from the camera and the environment, we found that fusing the output completion result with the original point cloud makes the performance significantly worse, which can be seen from the comparison of MSN (fusion) and MSN (vanilla). It also implies that our model is robust enough, which is conducive to rapid deployment in real robot experiments. Furthermore, compared with ideal results from the Oracle method, it demonstrates that point cloud completion remains an arduous task to solve.

\subsubsection{Ablation Studies} 
To comprehensively evaluate our proposed shape completion model, in this section we provide a series of ablation studies on our YCB-based dataset. Accordingly, the effectiveness of each special module in our model is analysed as follows:

\begin{table}[t!]
\caption{comparison of average EMD and CD from different point generators}
% \usepackage{multirow}
% \usepackage{graphicx}
\label{Generator}
\resizebox{0.48\textwidth}{!}{%
\begin{tabular}{c|ccc|ccc|l|l}
\hline
\multirow{2}{*}{\textbf{\begin{tabular}[c]{@{}c@{}}Similarity \\ Metrics\end{tabular}}} & \multicolumn{3}{l|}{\textbf{Uniform Distribution:}} & \multicolumn{3}{l|}{\textbf{Gaussian Distribution:}} & \multicolumn{2}{l}{\multirow{2}{*}{\textbf{ZERO}}} \\
                                                                                        & 0:1                  & -0.5:0.5        & -1:1       & 0.5,0.5/3        & 0,0.5       & 0,1                 & \multicolumn{2}{l}{}                               \\ \hline
Avg EMD                                                                                 & \textbf{5.94}        & 7.09            & 6.50       & 6.34             & 6.15        & \textbf{6.14}       & \multicolumn{2}{c}{9.88}                           \\
Avg CD                                                                                  & \textbf{1.89}        & 3.25            & 2.42       & 2.39             & 2.38        & \textbf{2.12}       & \multicolumn{2}{c}{6.17}                           \\ \hline
\end{tabular}%
}
\vspace{-1.5em}
\end{table}

\begin{table*}[htb!]
\centering
\caption{Influence of different surface numbers in the decoder}
\resizebox{1.0\textwidth}{!}{
\label{Surface}

\begin{tabular}{l|l|l|l|l}
\hline
\multirow{1}{*}{\textbf{Earth Mover's Distance (EMD)}} & \multirow{1}{*}{\textbf{n=4}} & \multirow{1}{*}{\textbf{n=8}} & \multirow{1}{*}{\textbf{n=16}} & \multirow{1}{*}{\textbf{n=32}}
\\\hline
Mug                                            & 4.71              & 3.94            & 3.70         & \textbf{3.61}                                        \\
Bleach cleanser                                           & 10.10              & 7.82          & 6.69         & {\textbf{5.94}}                                    \\ \hline
\end{tabular}

\begin{tabular}{l|l|l|l|l}
\hline
\multirow{1}{*}{\textbf{Chamfer Distance (CD)}} & \multirow{1}{*}{\textbf{n=4}} & \multirow{1}{*}{\textbf{n=8}} & \multirow{1}{*}{\textbf{n=16}} & \multirow{1}{*}{\textbf{n=32}}
\\\hline
Mug                                            & 9.01              & 6.70            &\textbf{6.61}         & 6.69                                        \\
Bleach cleanser                                           & 3.69              & 1.70          & \textbf{1.51}         & 1.53                                    \\ \hline
\end{tabular}}
\end{table*}
We first evaluate our transformer-based encoder module with other representative encoder modules under the same setting of convolutional/transformer layer number and object inputs. As shown in Tab.~\ref{Structure}, our encoder has a better result overall, though CMLP could get a great result on Mug's completion. When the point seed in the decoder is flat, we further analyze the influence of different point seed distributions and surface numbers in Tab.~\ref{Generator} and Tab.~\ref{Surface}. We can see that both Uniform and Gaussian sample method can achieve a better result at $(0, 1)$. We choose $Uniform(0,1)$ in our model, as it can achieve best results. Like the weight parameters in the neural network, the initialization value of points cannot be close to zero, which predicts the worst result. As illustrated in Tab.~\ref{Surface}, when the sub-surface number increases, the overall model performance improves. However, the improvement of completion results is limited when the number is above 16. 

\begin{table}[ht!]
\caption{Comparison of average difference of grasp joint and grasp pose from different completion type}
\label{simulation}
\resizebox{0.485\textwidth}{!}{%
\begin{tabular}{c|c|c|c|c|c|c}
\hline
\textbf{Error}  & \multicolumn{1}{l|}{\textbf{Partial}} & \multicolumn{1}{l|}{\textbf{Mirror}} & \multicolumn{1}{l|}{\textbf{Voxel-based}} & \multicolumn{1}{l|}{\textbf{RANSAC}} & \multicolumn{1}{l|}{\textbf{\begin{tabular}[c]{@{}c@{}}Ours\\  (canonical)\end{tabular}}} & \textbf{\begin{tabular}[c]{@{}c@{}}Ours\\  (arbitrary)\end{tabular}} \\ \hline
 \begin{tabular}[c]{@{}l@{}}Grasp Joint \\  (degree)\end{tabular} & 6.27                                   & 4.05                                 & 1.80                                      & 1.69                                  &\textbf{1.15}    
& 2.02        \\
\hline
\begin{tabular}[c]{@{}l@{}}Grasp Pose \\  (mm)\end{tabular}& 20.8                                  & 15.6                                 & 6.7                                       & 7.4                                  & 0.4                        &\textbf{0.2}
          \\ \hline
\end{tabular}}
\vspace{-1.5em}
\end{table}

\subsubsection{Visualization Analysis}
\begin{figure}[!t]
\includegraphics[width=0.48\textwidth]{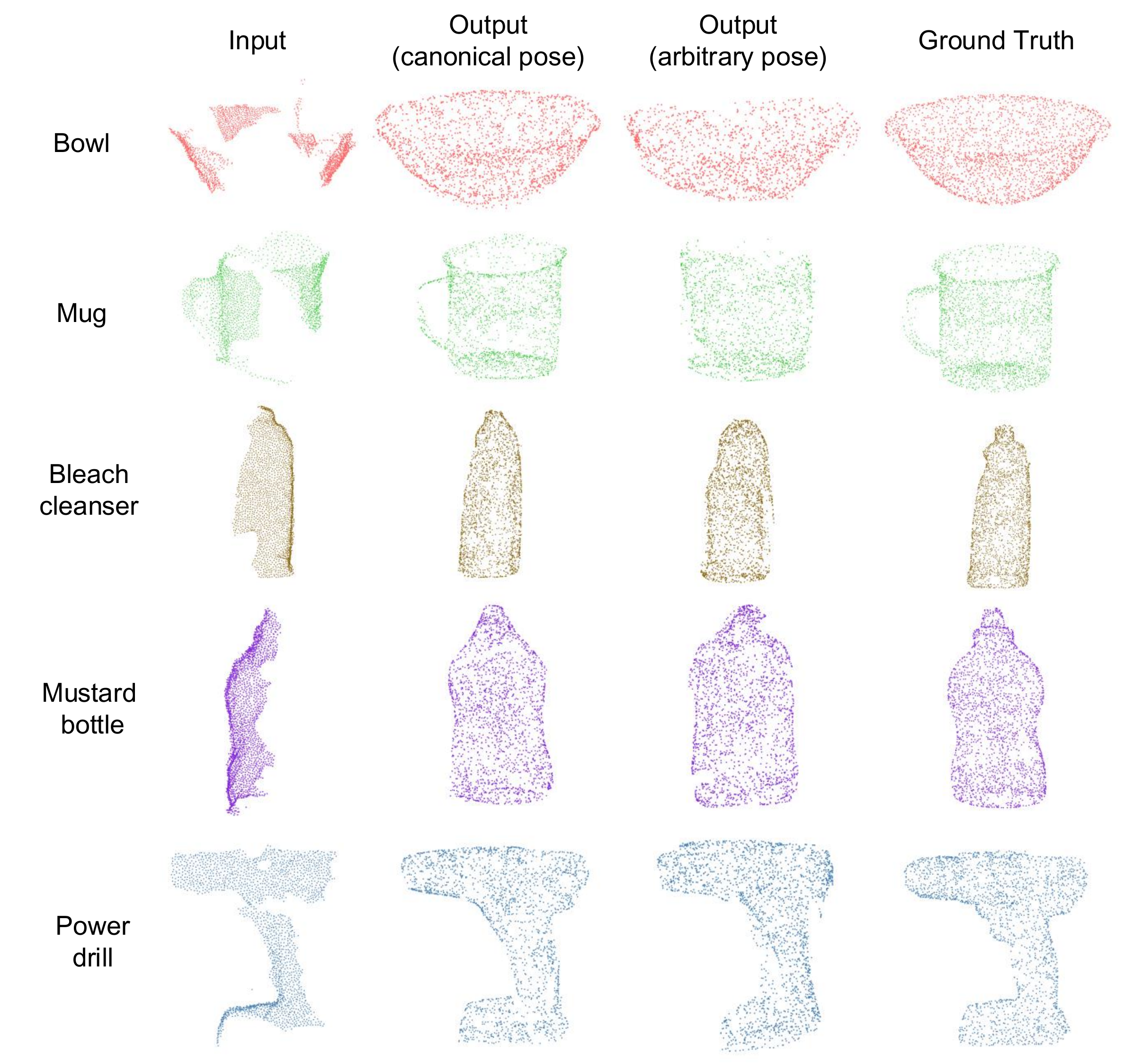}
\caption{Shape completion result using TransSC. The canonical pose result is trained under a fixed point cloud coordinate system while the arbitrary pose result is trained under the camera perspective. In the robot experiment, the arbitrary pose training result is used to generate grasps.}   
\label{Visulization}
\vspace{-0.5em}
\end{figure}

Fig.~\ref{Visulization} shows the visualized shape completion results using our TransSC. To facilitate visual analysis, the input partial point cloud of each object is first preprocessed to remove noisy data from the camera and the environment.  It can be seen that the geometric loss of the input point cloud in our dataset comes from the change of the camera viewpoint and the occlusion of other objects, which causes a big challenge for our model. The output results of the canonical pose show that our model works well on all simple and complex objects. Moreover, our model can generate realistic structures and details like the mug handle, bowl edge and bottle mouth. To enable robotic grasping, another shape completion model based on the arbitrary ground-truth pose is retrained through transforming the ground truth pose to the original pose of the input partial point cloud, and completion results are also shown in Fig.~\ref{Visulization}. Obviously, arbitrary output is not as good as the canonical output while it still restores the overall shape of each object well. It also demonstrates that achieving object completion of arbitrary poses in the real environment is still a formidable task. 

\subsection{Simulation Grasp Experiments with complete shapes}
We use Graspit!~\cite{miller2004graspit} to evaluate the quality of shape completion similar to~\cite{varley2017shape}. First, the Alpha shapes algorithm ~\cite{alpha_shape} is used to implement surface reconstruction of completion object. The output 3D mesh is then imported into GraspIt! Simulator to calculate grasps. To have a fair comparison, we also use Barrett Hand to generate grasps. After finishing the grasp generation, we remove the completion object and import the ground-truth object into the same place. Meanwhile, the Barrett Hand is moved back for 20 cm along the approach direction and then approaches the object until the gripper detect a collision or reach the calculated grasp pose. Furthermore, we adjust the gripper to the calculated grasp joint angles and perform the auto-grasp function in GraspIt! to ensure the gripper contacts with the object surface or reaches the joint limit. The joint angle difference and position difference are then recorded. We use four objects (bleach cleanser, cracker box, pitcher base and power drill) in the YCB objects set and calculate 100 grasps for each object in our experiment. 

We compare the average difference of joint angle and grasp pose from our shape completion model to that of Laplacian smoothing in Meshlab (Partial), mirroring completion~\cite{bohg2011mind} (Mirror), RANSAC-based approach~\cite{papazov2010efficient} and voxel-based completion~\cite{varley2017shape}. Note that we use two different models, canonical and arbitrary. The canonical model means all the training is transformed into the same object coordinate system and the arbitrary model means all the training data are transformed into the camera's coordinate system. Although from Fig.~\ref{Visulization} we can see the canonical model has a better shape completion result, but it requires a 6D pose of the target object if we want to map the complete point cloud into the real environment. To avoid this complication of adding a 6D pose estimation module and achieve real robot experiments, the arbitrary model is also trained. 
The simulation result is shown in Table.~\ref{simulation}.
It can be seen that Ours (canonical pose) gets the best simulation grasping performance, which outperforms other completion types. Ours (arbitrary pose) also obtains a great simulation result though its average joint pose is slightly bigger than RANSAC-based and voxel-based methods. Moreover, the average grasp pose errors of both models are smaller than other methods significantly. The larger joint error and lower pose error of Ours (arbitrary pose) indicates that corresponding completion object is slightly larger than the ground-truth object. The average difference from two models also demonstrates that  a perfect shape completion in an arbitrary pose is much harder than in a canonical pose.

\subsection{Robotic Experiments}
To evaluate the performance improvement using complete point cloud for robotic grasping, we choose six YCB objects to test the grasping success rate. The robot for evaluation is a UR5 robot arm equipped with a Robotiq 3-finger gripper. The vision sensor is an Industrial 3D camera from Mechmind~\footnote{https://en.mech-mind.net/} to acquire a high-quality partial point cloud. The selected six objects are list in Table.~\ref{tab:real_robot_experiment_result}. We select these objects because they are typical objects that may fail to generate good grasp candidates without shape completion. For other objects such as banana or marker, they are quite simple and small, which causes that improvement of shape completion on the grasping result is minor.

For the selected six objects, we perform grasp evaluation on two different methods: PointNetGPD grasp with/without shape completion. We run the robot experiment by randomly putting the object on the table and grasping for ten times, then calculating the success rate. The experiment result is shown in Table.~\ref{tab:real_robot_experiment_result}. We can see that all six objects' grasp success rates using PointNetGPD with TransSC outperform or even with original method. 
The low success rate of power drill for both methods is because when the robot tries to grasp the head of the power drill, the contact area is too slippy. The failures of PointNetGPD with observed point cloud input are mainly from the limit of camera viewpoint, and GPG generates grasp candidates that sink into the object. An example of this situation is shown in Fig.~\ref{fig:grasp_candidate}. This is a strong evidence that our shape completion model can improve the grasp success rate in some particular objects. 

\begin{table}[t!]
\caption{Real robot experiment result}
\label{tab:real_robot_experiment_result}
\resizebox{0.48\textwidth}{!}{%
\begin{tabular}{c|ccccccc}
\hline
\textbf{Method}                                                     & \textbf{\begin{tabular}[c]{@{}c@{}}cracker\\  box\end{tabular}} & \textbf{mug} & \textbf{\begin{tabular}[c]{@{}c@{}}meat\\ can\end{tabular}} & \textbf{\begin{tabular}[c]{@{}c@{}}pitcher\\ base\end{tabular}} & \textbf{\begin{tabular}[c]{@{}c@{}}bleach\\ cleanser\end{tabular}} & \textbf{\begin{tabular}[c]{@{}c@{}}power\\ drill\end{tabular}} & \textbf{average} \\ \hline
PointNetGPD\cite{liang2019pointnetgpd}                                                         & 70\%                                                            & 70\%         & 80\%                                                        & 80\%                                                            & 90\%                                                               & 40\%                                                           & 71.67\%          \\ \hline
\begin{tabular}[c]{@{}c@{}}PointNetGPD \\ with TransSC\end{tabular} & 80\%                                                            & 100\%        & 100\%                                                       & 80\%                                                            & 90\%                                                               & 50\%                                                           & 83.33\%          \\ \hline
\end{tabular}}
\vspace{-1.5em}
\end{table}
\section{Conclusion and Future Work}
We present a novel transformer-based shape completion network (TransSC), which is robust to sparse and noisy point cloud input. A transformer-based encoder and manifold-based decoder are designed in our network, which makes our model achieve a great completion result and outperform other representative methods. Besides, TransSC could be easily embedded into a grasp evaluation pipeline and improve grasping performance significantly.

The lack of geometric information on the object in our dataset is not only due to the change of the camera viewpoint but also the occlusion of different objects. Thus, TransSC could also achieve shape completion for occluded objects. 
In future work, our goal is to integrate semantic segmentation into our shape completion pipeline to make the robot grasp objects better in a cluttered environment.

\section{Acknowledgement}
This research was funded by the German Research Foundation (DFG)
and the National Science Foundation of China (NSFC)
in project Crossmodal Learning, DFG TRR-169/NSFC 61621136008,
and partially supported by European projects H2020 STEP2DYNA (691154) and Ultracept (778602).
We also thanks Mech-Mind Robotics Company for providing the 3D camera.
\addtolength{\textheight}{-4cm}  

\bibliographystyle{IEEEtran}
\bibliography{bibtex/bib/reference}

\end{document}